\documentclass{article}
\usepackage{spconf,amsmath,graphicx}
\usepackage[colorlinks=true]{hyperref}
\usepackage{multirow}
\usepackage{subcaption}
\usepackage[T1]{fontenc}
\usepackage{cite}
\usepackage{booktabs}
\usepackage{amsmath}
\usepackage{marvosym}
\usepackage{subcaption}
\usepackage{float}
\usepackage[marginal]{footmisc}

\usepackage{amssymb}
\usepackage{float}

\title{Multimodal Multi-Agent Empowered Legal Judgment Prediction}
\name{
\begin{tabular}{c}
\vspace*{-0.16em}
Zhaolu Kang\textsuperscript{1,2,*}, Junhao Gong\textsuperscript{1,*}, Qingxi Chen\textsuperscript{1}, Hao Zhang\textsuperscript{1}, Jiaxin Liu\textsuperscript{3} \\
Rong Fu\textsuperscript{4}, Zhiyuan Feng\textsuperscript{5}, Yuan Wang\textsuperscript{6}, Simon Fong\textsuperscript{4}, Kaiyue Zhou\textsuperscript{2,\Letter}\thanks{\textsuperscript{*}Equal Contribution. \textsuperscript{\Letter}Corresponding author.
}
\thanks{$^{1}$ JurisMM is available at: \href{https://github.com/Zhaolu-K/JurisMMA}{https://github.com/Zhaolu-K/JurisMMA}.}
\thanks{$^{2}$This study was supported in part by Chengdu Key Research and Development Program (No. 2025-XT00-00005-GX).}
\end{tabular}
}
\address{
\vspace*{-0.17em}
\textsuperscript{1}Peking University
\textsuperscript{2}Chengdu Minto Tech
\textsuperscript{3}University of Illinois Urbana-Champaign\\
\textsuperscript{4}University of Macau
\textsuperscript{5}Tsinghua university
\textsuperscript{6}Zhejiang University
\vspace*{-0.17em}
\\\texttt{zlkang25@stu.pku.edu.cn,}
\texttt{kyzhou@wayne.edu}
}

\begin{document}
%
%
\maketitle

\begin{figure*}[!t]
    \centering
    \includegraphics[width=1\linewidth]{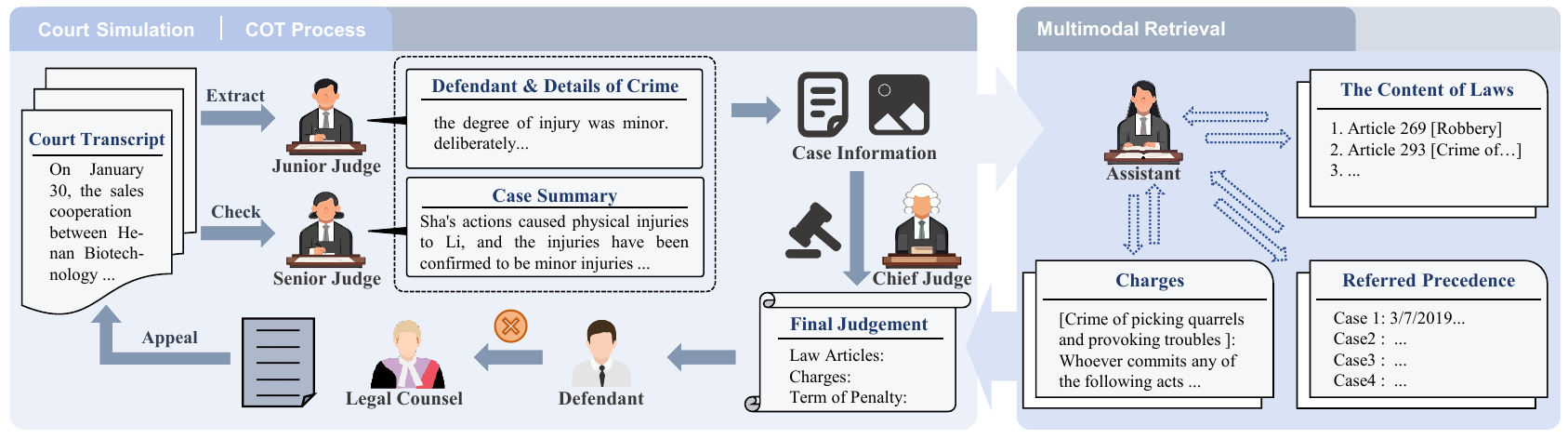}
    \vspace{-20pt}
    \caption{Overview of our framework \textit{JurisMMA}.}
    \vspace{-10pt}
    \label{main}
\end{figure*}

\begin{abstract}
Legal Judgment Prediction (LJP) aims to predict the outcomes of legal cases based on factual descriptions, serving as a fundamental task to advance the development of legal systems. 
Traditional methods often rely on statistical analyses or role-based simulations but face challenges with multiple allegations, diverse evidence, and lack adaptability. 
In this paper, we introduce \textit{\textbf{JurisMMA}}, a novel framework for LJP that effectively decomposes trial tasks, standardizes processes, and organizes them into distinct stages. 
Furthermore, we build \textit{\textbf{JurisMM}}\footnotemark[1], a large dataset with over 100,000 recent Chinese judicial records, including both text and multimodal video-text data, enabling comprehensive evaluation. 
Experiments on \textit{JurisMM} and the benchmark LawBench validate our framework’s effectiveness. 
These results indicate that our framework is effective not only for LJP but also for a broader range of legal applications, offering new perspectives for the development of future legal methods and datasets.\footnotemark[2]
\end{abstract}
\begin{keywords}
Legal Judgment Prediction, Multi-Agent Framework, MLLMs
\end{keywords}
\section{Introduction}
\label{sec:intro}


Natural Language Processing (NLP) shows great promise in the legal field, with Legal Judgment Prediction (LJP) \cite{xu2020distinguish,malik2021ildc,wei2026fademembiologicallyinspiredforgettingefficient} emerging as a core task. LJP predicts court outcomes (e.g., applicable laws, charges, penalties) from case facts, improving judicial efficiency and providing public legal guidance. State-of-the-art Neural Networks \cite{2018TOPJUDGE,mpbfn2019} and Large Language Models (LLMs) \cite{lin2023pushing,zhao2025thinking,kang2025jurisctc,zhusymphony,li2025graph,lin2024adaptsfl} have yielded promising results.

However, the unique characteristics of legal language, including its high degree of specialization, complexity, and reliance on precise terminology, pose significant challenges for LJP systems.
These challenges are further compounded by the growing global population, which has led to an increase in the number of judicial cases, placing immense pressure on the legal systems.
Efforts to mitigate these problems have focused on simplifying the simulation of real courtrooms \cite{zhang2025pdtrim,chen2024agentcourtsimulatingcourtadversarial}, utilizing Retrieval-Augmented Generation (RAG) techniques \cite{he-etal-2024-agentscourt,LegalReasoner2024,wei2026thinkaugmentedfunctioncallingimproving}, and refining prompt engineering \cite{zhang2025trimtokenator,xie2025chat}. Although neural network classifiers have demonstrated effectiveness in straightforward prediction tasks, they often lack the capacity to model the complex procedural and interactive aspects of real-world trials.
In summary, several challenges remain in the LJP task:

1) \textit{Real-Scenario Reproduction}: Most frameworks fail to predict multiple charges/laws or utilize multimodal evidence (e.g., trial recordings), hindering outcome accuracy.

2) \textit{Multivariable Interactions}: Simplistic role-based simulations ignore shared agent functions and procedural design, leading to one-sided conclusions and illogical trial outcomes.

To address these challenges, we propose \textit{JurisMMA}, a multi-agent LJP framework illustrated in Figure \ref{main}, which leverages the reasoning capabilities of LLMs to simulate court proceedings. The framework consists of two major components: (1) a six-agent courtroom module involving Junior Judges, Senior Judges, Assistants, the Chief Judge, Legal Counsel, and Defendants, which is designed to follow standard trial procedures and allows dissatisfied Defendants to file an appeal; and (2) a knowledge retrieval module in which an Assistant provides access to the most recent rulings and legal statutes.
In addition, we introduce \textit{JurisMM}, a dataset comprising over 100,000 up-to-date real cases collected from off
icial sources, along with 83 multimodal samples, aimed at addressing outdating problem in existing benchmarks.
We summarize our contributions as follows:
\vspace{-0.3cm}
\begin{itemize}
    \item  \textit{JurisMMA} is a new multi-agent framework capable of conducting comprehensive simulations of court trials based on the provided case information.
    \vspace{-0.3cm}
    \item We propose a novel dataset \textit{JurisMM}  comprising single-text modality and multimodal video-text data, consisting of more than 100,000 judicial cases based on the latest criminal law judgments.
   \vspace{-0.3cm}
    \item Extensive experiments and ablation studies demonstrate that, across all aspects of several key LJP tasks including Law Articles prediction, Charges prediction, and Terms of Penalty prediction, our framework outperforms existing state-of-the-art methods.
   \vspace{-0.3cm}
\end{itemize}

\section{Method}
\label{sec:format}

To uphold the standard logic and operational norms of judicial adjudication, we design a structured six-stage workflow in the proposed \textit{JurisMMA}, which incorporates six distinct agent roles: Junior Judge (JJ), Senior Judge (SJ), Chief Judge (CJ), Assistant, Defendant, and Legal Counsel. These roles are assigned to different stages to collaborate and complete the entire adjudication cycle.

\subsection{Stage I: Information Extraction}
This phase initiates first-instance proceedings, relying on the collaboration between the Junior Judge and the Senior Judge. Both agents receive the complete criminal case text \(T \in \mathcal{T}\):
The Junior Judge first extracts core factual elements: 
\vspace{-7pt}
\begin{multline}
E = \{ e_1, e_2, e_3 \} = \{\text{defendant identification}, \\
\text{crime details}, \text{criminal motive}\}.
\end{multline}

to generate structured judicial facts:
 \vspace{-6pt}
  \begin{equation}
  F_{JJ} = \mathrm{Extract}(T, E) = \{ f_{e_1}, f_{e_2}, f_{e_3} \},
  \end{equation}
and the Senior Judge then integrates \(F_{JJ}\) with the full case narrative to produce a concise Case Summary using a pretrained text encoder \(\mathrm{Encode}(\cdot)\) and a aggregation function:
\vspace{-7pt}
\begin{equation}
S = \text{Aggregate}(\text{Check}(\text{Encode}(T), \text{Encode}(F_{JJ})), T,F_{JJ}),
\end{equation}
The Senior Judge examines the consistency of facts between \(F_{JJ}\) and \(T\), simultaneously marks implicit associative information that needs to be supplemented, such as timelines and descriptions of chains of evidence, and ultimately produces a standardized case summary.

\subsection{Stage II: Legal Information Retrieval}
The Assistant acts as an auxiliary agent, using the Case Summary \(S\) to perform multi-source retrieval across three legal knowledge bases: statutes \(D_{\mathrm{law}}\), charge definitions \(D_{\mathrm{charge}}\), and precedents \(D_{\mathrm{precedent}}\).

For \(D_{\mathrm{law}}\) and \(D_{\mathrm{charge}}\), we use dense vector retrieval via FAISS, 
with similarity calculated using cosine similarity between \(S\)’s embedding 
\(\mathbf{v}_S = \mathrm{Encode}(S)\) and document embeddings \(\mathbf{v}_d\).
The top-\(k\) documents are selected:
\vspace{-3pt}
\begin{equation}
\begin{cases}
R_{\text{law}} = \mathrm{TopK}(\mathrm{Sim}(S, D_{\text{law}}), 10), \\
R_{\text{charge}} = \mathrm{TopK}(\mathrm{Sim}(S, D_{\text{charge}}), 10).
\end{cases}
\end{equation}

For \(D_{\mathrm{precedent}}\), hybrid retrieval is performed by combining BM25 and cosine similarity:
  \begin{equation}
  \mathrm{Score}(S, p) = \alpha \cdot \mathrm{BM25}(S, p) + (1-\alpha) \cdot \mathrm{Sim}(S, p),
  \end{equation}
  where \(\alpha \in [0,1]\) balances sparse and dense scoring.

All results are consolidated into a unified Case Information Document:
\vspace{-3pt}
  \begin{equation}
  I = \mathrm{Concat}(R_{\mathrm{law}}, R_{\mathrm{charge}}, R_{\mathrm{precedent}}).
  \end{equation}

\subsection{Stage III: First-instance Decision}
Adjudicatory authority shifts to the Chief Judge, who synthesizes JJ’s facts \(F_{JJ}\), SJ’s summary \(S\), and Assistant’s retrieval results \(I\) to issue the first-instance judgment. Charges are selected by aligning candidate charges with facts and precedents:
\begin{equation}
\begin{split}
C = \arg\max_{c \in R_{\mathrm{charge}}}[ \ 
& \beta \cdot \mathrm{Align}(c, F_{JJ}) \\
& + (1-\beta) \cdot \mathrm{Align}(c, R_{\mathrm{precedent}}) ],
\end{split}
\end{equation}
where \(\mathrm{Align}(c, F_{JJ})\) normalizes cosine similarity of charge \(c\) with \(F_{JJ}\), and \(\mathrm{Align}(c, R_{\mathrm{precedent}})\) averages cosine similarities between \(c\) and charges of retrieved precedents.

Sentencing Calculation means Sentence L (in months) adjusts the statutory base sentence Base(C) by taking into account precedent influence and mitigating factors.
  \begin{equation}
  \begin{cases}
  D = \sum_{p \in R_{\mathrm{precedent}}} \gamma_p \cdot \mathrm{Devi}(p, F_{JJ}), \\
  L = \mathrm{Base}(C) + D \delta \cdot M,
  \end{cases}
  \end{equation}
  where \(\gamma_p\) weights precedent \(p\)’s influence, \(\mathrm{Devi}(p, F_{JJ})\) measures factual deviations, \(M\) is the number of mitigating factors, and \(\delta > 0\) is a reduction coefficient.

Relevant laws exceeding a threshold \(\theta\) are filtered (\(R_{\mathrm{law}}^* = \{ l \in R_{\mathrm{law}} \mid \mathrm{Match}(l, F_{JJ}) \geq \theta \}\)), and the judgment is finalized as:
\vspace{-3pt}
  \begin{equation}
  J = \{ R_{\mathrm{law}}^*, C, L \}.
  \end{equation}

\subsection{Stage IV: Defendant’s Response}
The Defendant may accept \(J\), which concludes the case, or raise valid objections. Objections are defined as claims that diverge from \(J\) and are supported by the original case text \(T\):
\vspace{-6pt}
\begin{equation}
O = \{ o \mid \mathrm{Diff}(o, J) > \tau \text{ and } \mathrm{Sup}(o, T) > \tau \},
\end{equation}
where \(\mathrm{Diff}(o, J)\) quantifies semantic/factual divergence between \(o\) and \(J\), \(\mathrm{Sup}(o, T)\) measures the supportiveness of the original case text \(T\) for objection \(o\) through textual entailment, and \(\tau \in [0,1]\) is the validity threshold. The objection set \(O\) forms the basis for potential appeals.

\subsection{Stage V: Appeal Submission}
If \(O \neq \emptyset\), the Legal Counsel constructs an appeal brief by integrating \(O\) with supporting laws and contextual reasoning:
\vspace{-6pt}
\begin{equation}
A = \bigcup_{o \in O} \left\{ o \cup \{ l \in R_{\mathrm{law}} \mid \mathrm{Fv}(l, o) > \eta \} \cup \mathrm{A}(o, S) \right\},
\end{equation}
where \(\mathrm{Fv}(l, o)\) measures legal support of law \(l\) for objection \(o\), \(\eta \in [0,1]\) is the support threshold, and \(\mathrm{A}(o, S)\) denotes contextual reasoning (using \(S\) to strengthen arguments). The appeal brief \(A\) triggers the second-instance process.

\subsection{Stage VI: Second-instance Review}
First, the validity of the appeal brief \(A\) is scored:
\vspace{-6pt}
\begin{equation}
\mathrm{Val}(A) = \frac{1}{|A|} \sum_{a \in A} [ \theta \cdot \mathrm{Valid}(a, T) + (1-\theta) \cdot \mathrm{Valid}(a, R_{\mathrm{law}}) ],
\end{equation}
where \(\theta \in [0,1]\) balances factual validity from \(T\) and legal validity from \(R_{\mathrm{law}}\), and \(\mathrm{Valid}(\cdot, \cdot)\) quantifies argument soundness. 
The final judgment is then determined:
\vspace{-6pt}
\begin{equation}
J' = 
\begin{cases}
\mathrm{Revise}(J, A), & \text{if } \mathrm{Val}(A) \geq \epsilon, \\
J, & \text{otherwise},
\end{cases}
\end{equation}
where \(\epsilon \in [0,1]\) is the appeal acceptance threshold. 
\(\mathrm{Revise}(J, \\ A)\) indicates that CJ will take \(J\) and \(A\) as supplementary inputs, re-execute Stages I and II, and render a final judgment based on the summarized information.
Under the two-instance trial system, \(J'\) is final, marking the end of the adjudication cycle.

\section{Experiments and Results}
\label{sec:pagestyle}


We construct the JurisMM multimodal dataset, which consists of text and video subsets, to support legal judgment prediction (LJP). To address the limitations of outdated legal text datasets like CAIL2018, we collected 150,000 public criminal judgments from 2014 to early 2024 on China Judgments Online, covering all types of criminal offenses. By excluding cases involving articles amended between 2014-2025 to ensure compliance with the Criminal Law of the People's Republic of China (2023 Amendment) and balancing the category distribution, we finalized the JurisMM-Text subset with 101,544 cases, split into 94,494 for training and 7,050 for testing.

To supplement the multimodal legal data, we develop a semi-automatic framework to build the JurisMM-Video subset. First, we segment videos and use vision-language models to filter clips related to crime scenes, evidence, or suspect actions. Then, legal experts have manually reviewed the clips, selected key frames, and formulated questions. This process yields 83 high-quality multimodal samples that capture critical visual information, such as evidence, scene topography, and suspect behavior, for context-aware LJP.

In addition to the dataset, we have built a domain-specific knowledge repository that draws on the 2023 Criminal Law Amendment. It has three core components: Legal Content, which includes 438 effective criminal law articles from the National Laws and Regulations Database 2023, supplemented with Supreme People's Court judicial interpretations; Charges, a structured list of 483 major charges for prediction and retrieval; and Precedents, which consists of 46,365 representative cases from JurisMM-Text to aid legal reasoning.
\begin{table*}[htbp]
\vspace{-5pt}
\centering
\setlength{\tabcolsep}{5pt} 
\renewcommand{\arraystretch}{0.45} 
\resizebox{\linewidth}{!}{
\begin{tabular}{c|cccc|cccc|ccccc}
\toprule
\multirow{2}{*}{\textbf{Model}} & \multicolumn{4}{c|}{\textbf{Law Articles}} & \multicolumn{4}{c|}{\textbf{Charges}} & \multicolumn{5}{c}{\textbf{Terms of Penalty}} \\
\cmidrule(lr){2-5} \cmidrule(lr){6-9} \cmidrule(lr){10-14}
 & \textbf{Acc.} & \textbf{MP} & \textbf{MR} & \textbf{MF} & \textbf{Acc.} & \textbf{MP} & \textbf{MR} & \textbf{MF} & \textbf{Acc.} & \textbf{MP} & \textbf{MR} & \textbf{MF} & \textbf{N-Id} \\
\midrule
TextCNN & 0.703 & 0.389 & 0.343 & 0.334 & 0.696 & 0.492 & 0.457 & 0.441 & 0.325 & 0.321 & 0.282 & 0.285 & --- \\
TOPJUDGE & 0.768 & 0.455 & 0.411 & 0.400 & 0.751 & 0.516 & 0.519 & 0.491 & 0.373 & 0.358 & 0.364 & 0.342 & --- \\
MPBFN & 0.755 & 0.467 & 0.391 & 0.391 & 0.707 & 0.427 & 0.416 & 0.387 & 0.324 & 0.343 & 0.267 & 0.263 & --- \\
\midrule
GLM-4V-9B & 0.095 & 0.019 & 0.027 & 0.015 & 0.476 & 0.169 & 0.120 & 0.119 & 0.199 & 0.320 & 0.197 & 0.163 & 0.668 \\
mPLUG-7B & 0.094 & 0.027 & 0.035 & 0.019 & 0.319 & 0.093 & 0.060 & 0.058 & 0.100 & 0.250 & 0.168 & 0.067 & 0.507 \\
Qwen2.5-VL-3B-Instruct & 0.220 & 0.089 & 0.072 & 0.057 & 0.258 & 0.109 & 0.068 & 0.068 & 0.193 & 0.169 & 0.122 & 0.100 & 0.787 \\
Qwen2.5-VL-7B-Instruct & 0.271 & 0.133 & 0.121 & 0.092 & 0.134 & 0.061 & 0.021 & 0.026 & 0.188 & 0.135 & 0.119 & 0.096 & 0.799 \\
Qwen2.5-7B-Instruct & 0.523 & 0.133 & 0.170 & 0.123 & 0.590 & 0.161 & 0.144 & 0.134 & 0.237 & 0.227 & 0.145 & 0.124 & 0.826 \\
GPT-4o-20250326 & 0.380 & 0.264 & 0.207 & 0.186 & 0.436 & 0.253 & 0.186 & 0.182 & 0.372 & 0.432 & 0.357 & 0.319 & 0.842 \\
\midrule
Qwen2.5-VL-7B(SFT) & 0.365 & 0.207 & 0.241 & 0.200 & 0.515 & 0.295 & 0.277 & 0.252 & 0.252 & 0.123 & 0.096 & 0.040 & 0.818 \\
\midrule
\textit{JurisMMA (w GPT-4o)} & \textbf{0.872} & \textbf{0.604} & \textbf{0.645} & \textbf{0.602} & \textbf{0.882} & \textbf{0.600} & \textbf{0.642} & \textbf{0.603} & \textbf{0.409} & \textbf{0.520} & \textbf{0.393} & \textbf{0.361} & \textbf{0.848} \\
\quad w/o KB & 0.653 & 0.323 & 0.323 & 0.281 & 0.801 & 0.203 & 0.190 & 0.187 & 0.378 & 0.454 & 0.371 & 0.335 & \textbf{0.848} \\
\quad w/o MA & 0.767 & 0.592 & 0.607 & 0.569 & 0.773 & 0.581 & 0.599 & 0.566 & 0.303 & 0.475 & 0.304 & 0.272 & 0.803 \\
\midrule
\textit{JurisMMA (w Qwen2.5-VL-7B)} & 0.624 & 0.268 & 0.313 & 0.264 & 0.618 & 0.384 & 0.424 & 0.370 & 0.189 & 0.175 & 0.123 & 0.103 & 0.784 \\
\quad w/o KB & 0.389 & 0.159 & 0.143 & 0.132 & 0.414 & 0.191 & 0.116 & 0.129 & 0.216 & 0.258 & 0.148 & 0.132 & 0.782 \\
\quad w/o MA & 0.579 & 0.254 & 0.293 & 0.253 & 0.604 & 0.341 & 0.382 & 0.336 & 0.207 & 0.255 & 0.149 & 0.125 & 0.789 \\
\bottomrule
\end{tabular}
}
\vspace{-10pt}
\caption{Overall performance and ablation results on \textit{JurisMM-Text}.}
\label{tab:resultall}
\vspace{-10pt}
\end{table*}

\subsection{Baselines}
For traditional neural networks, we select representative and competitive baselines TextCNN, TOPJUDGE~\cite{2018TOPJUDGE} and MPBFN~\cite{mpbfn2019} to benchmark our framework; despite their age, these models remain competitive on \textit{JurisMM} due to the lack of recent breakthroughs in legal-specific neural architectures. For LLM-based baselines, we evaluate open-source models including GLM-4V-9B~\cite{glm2024chatglmfamilylargelanguage}, mPLUG-Owl-7B~\cite{ye2024mplugowlmodularizationempowerslarge}, Qwen2.5-VL-3B/7B-Instruct~\cite{bai2025qwen25vltechnicalreport}, Qwen2.5-7B-Instruct~\cite{qwen2025qwen25technicalreport}, as well as the closed-source GPT-4o-20250326~\cite{openai2024gpt4ocard}. Since no open-source LLM is pre-trained specifically for LJP, we fine-tune Qwen2.5-VL-7B-Instruct on our legal dataset to provide a more relevant comparison.

\subsection{Experimental Settings}
Neural network models are trained on the \textit{JurisMM-Text} training split and evaluated on the test set, using the Adam optimizer with a learning rate of \(10^{-3}\), a dropout rate of 0.5, a batch size of 512, and a training duration of 16 epochs. For LLMs, closed-source models are evaluated with the temperature fixed at 0 to ensure deterministic outputs, while open-source models are run three times and average results are reported; \textit{JurisMMA} primarily uses GPT-4o, with additional experiments using Qwen2.5-VL-7B to test generalizability, and token consumption averages around 20,000 tokens per input (varying with case complexity). Performance is evaluated using four widely adopted multi-classification metrics: accuracy (Acc.), macro-precision (MP), macro-recall (MR), and macro-F1 (F1); for the sentence length prediction task, we discretize continuous penalty durations into non-overlapping intervals and evaluate performance using the Normalized Log-Distance (N-Ld) metric from LawBench~\cite{fei2023lawbench}.

\subsection{Overall Performance}

Table~\ref{tab:resultall} details the comparison of all evaluated models on the \textit{JurisMM-Text} dataset across three core legal judgment tasks: Law Articles prediction, Charges prediction, and Terms of Penalty prediction. Our proposed framework \textit{JurisMMA} consistently outperforms all baselines in every task and metric.
\begin{figure}[!t]
    \centering
    \vspace{-10pt}
    \includegraphics[width=1\linewidth]{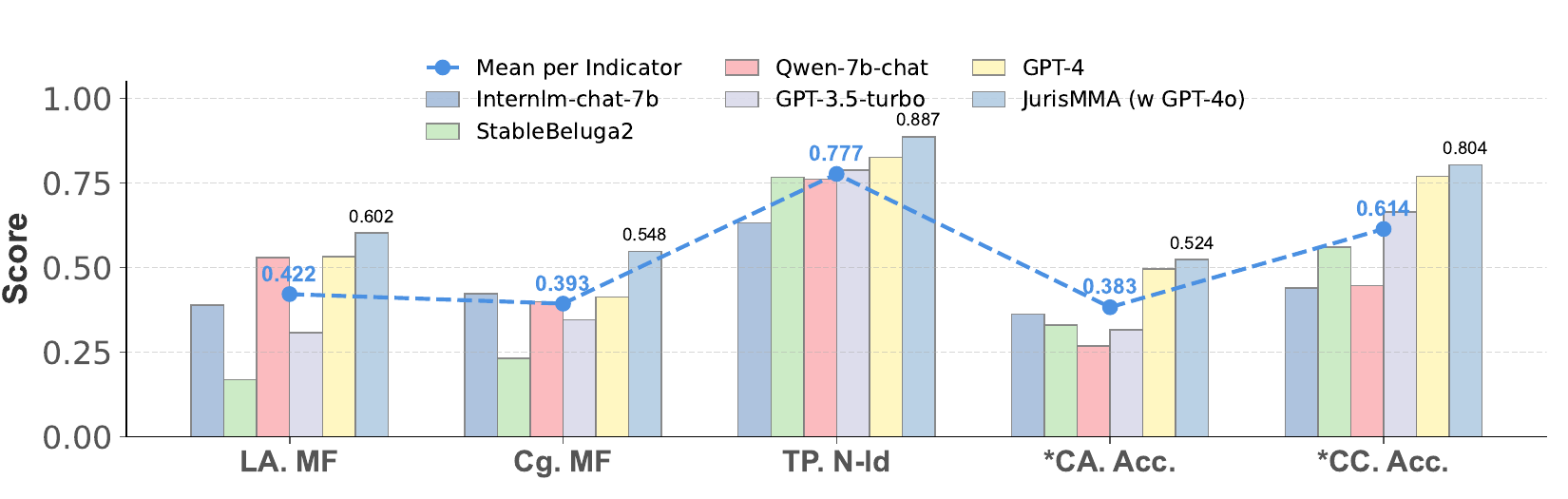}
    \vspace{-20pt}
    \caption{Overall performance on LawBench.}
    \label{fig:lawbench}
    \vspace{-20pt}
\end{figure}
All models perform best in Law Articles prediction, followed by Charges prediction, while Terms of Penalty prediction is the most challenging. Though GPT-4o leads among LLMs, it still lags behind neural networks in most tasks, indicating LLMs’ general language skills are insufficient for specialized legal judgment without fine-tuning. Fine-tuning the Qwen2.5-VL-7B model on the legal dataset significantly improves its performance compared to the base version, yet it still underperforms \textit{JurisMMA} (integrated with Qwen2.5-VL-7B) in most evaluation metrics.


We further evaluate \textit{JurisMMA} on LawBench, a comprehensive Chinese legal reasoning benchmark, selecting five representative tasks (three aligned with its design, plus prevalent Case Analysis (CA) and Crime Amount Calculation (CC)). As shown in Figure~\ref{fig:lawbench}, \textit{JurisMMA} outperforms five general-purpose LLMs on all tasks, verifying the adaptability of its legal-tailored multi-agent framework.
\begin{table}[!t]
\centering
\setlength{\tabcolsep}{7pt} 
\renewcommand{\arraystretch}{0.45} 
\resizebox{\columnwidth}{!}{
\begin{tabular}{c|cc|cc|cc}
\toprule
\multirow{2}{*}{\textbf{Model}} & \multicolumn{2}{c|}{\textbf{Law Articles}} & \multicolumn{2}{c|}{\textbf{Charges}} & \multicolumn{2}{c}{\textbf{Terms of Penalty}} \\
\cmidrule(lr){2-3} \cmidrule(lr){4-5} \cmidrule(lr){6-7}
 & \textbf{Acc.} & \textbf{MF} & \textbf{Acc.} & \textbf{MF} & \textbf{N-Id} & \textbf{MF} \\
\midrule
GPT-4o & 0.518  & 0.165 & 0.325 & 0.183 & 0.724 & 0.114 \\
\quad w picture & 0.518 & 0.180 & 0.337 & 0.177 & 0.779 & 0.164 \\
\midrule
\textit{JurisMMA}& 0.614 & 0.334 & 0.518 & 0.309 & 0.744 & 0.105 \\
\quad w picture &  \textbf{0.639} & \textbf{0.336} &  \textbf{0.554} &  \textbf{0.353} & \textbf{0.816} & \textbf{0.184} \\

\bottomrule
\end{tabular}
}
\vspace{-10pt}
\caption{Overall performance on \textit{JurisMM-Video}.}\label{tab:resultimage}
\vspace{-14pt}
\end{table}

We also assess \textit{JurisMMA} on 83 multimodal legal cases from \textit{JurisMM-Video}, with our evaluation results presented in Table~\ref{tab:resultimage}. 
Incorporating visual information consistently boosts its accuracy and macro-F1 across tasks by enriching contextual understanding. 

To systematically evaluate the impact of key components in \textit{JurisMMA}, ablation experiments were conducted on the \textit{JurisMM-Text} test set. Table~\ref{tab:resultall} presents results of removing the knowledge base (KB) and multi-agent collaboration (MA).
Removing the knowledge base (w/o KB) and multi-agent collaboration (w/o MA) both cause significant performance drops, confirming their crucial roles. GPT-4o outperforms Qwen2.5-VL-7B overall, but the impact of ablating KB or MA is consistent across models, showing these components are vital regardless of backbone.

\section{Conclusion}
\label{sec:typestyle}

We propose \textit{JurisMMA}, a multi-agent framework for alleviating the inherent complexities of LJP by decomposing trial tasks into structured stages and simulating realistic courtroom procedures. 
By leveraging the reasoning capabilities of LLMs within a carefully orchestrated multi-agent system, our method effectively captures the multivariable interactions and procedural nuances that characterize real-world legal cases. To further support this endeavor, we introduce \textit{JurisMM}, a large-scale, up-to-date dataset comprising over 100,000 judicial cases, including both unimodal textual data and multimodal video-text samples.

\vfill\pagebreak

\bibliographystyle{IEEEbib}
\bibliography{strings,refs}

\end{document}